\documentclass[conference]{IEEEtran}
\usepackage{multirow}

\IEEEoverridecommandlockouts

\usepackage{cite}
\usepackage{amsmath,amssymb,amsfonts}
\usepackage{algorithmic}
\usepackage{graphicx}
\usepackage{textcomp}
\usepackage{booktabs}
\usepackage{hyperref}
\usepackage[table]{xcolor}  
\def\BibTeX{{\rm B\kern-.05em{\sc i\kern-.025em b}\kern-.08em
    T\kern-.1667em\lower.7ex\hbox{E}\kern-.125emX}}

\usepackage{tikz}
\newcommand\copyrighttext{%
	\footnotesize Accepted at ITSC 2025, \copyright IEEE. Personal use is permitted, but republication/redistribution requires IEEE permission.  Permission from IEEE must be obtained for all other uses, in any current or future media, including reprinting/republishing this material for advertising or promotional purposes, creating new collective works, for resale or redistribution to servers or lists, or reuse of any copyrighted component of this work in other works.}
\newcommand\copyrightnotice{%
	\begin{tikzpicture}[remember picture,overlay]
	\node[anchor=south,yshift=10pt] at (current page.south) {\fbox{\parbox{\dimexpr\textwidth-\fboxsep-\fboxrule\relax}{\copyrighttext}}};
	\end{tikzpicture}%
}

\begin{document}

\title{
\textsc{T-Mask}: Temporal Masking for Probing Foundation Models  across Camera Views in Driver Monitoring %
}

\author{\IEEEauthorblockN{Thinesh Thiyakesan Ponbagavathi}
\IEEEauthorblockA{\textit{Institute of Artificial Intelligence} \\
\textit{University of Stuttgart}\\
Stuttgart, Germany
}
\and
\IEEEauthorblockN{Kunyu Peng}
\IEEEauthorblockA{\textit{Institute for Anthropomatics and Robotics } \\
\textit{Karlsruhe Institute of Technology}\\
Karlsruhe, Germany 
}
\and
\IEEEauthorblockN{Alina Roitberg}
\IEEEauthorblockA{\textit{Institute of Artificial Intelligence} \\
\textit{University of Stuttgart}\\
Stuttgart, Germany 
}
}

\maketitle
\copyrightnotice{}
\thispagestyle{empty}
\pagestyle{empty}

\begin{abstract}

Changes of camera perspective are a common obstacle in driver monitoring. 
 While deep learning and pretrained foundation models show strong potential for improved generalization via lightweight adaptation of the final layers (``probing’’), their robustness to unseen viewpoints remains underexplored.
We study this challenge by adapting image foundation models to driver monitoring using a single training view, and evaluating them directly on unseen perspectives without further adaptation.
We benchmark simple linear probes, advanced probing strategies, and compare two foundation models (DINOv2 and CLIP) against parameter-efficient fine-tuning (PEFT) and full fine-tuning.

Building on these insights, we introduce \textsc{T-Mask} -- a new image-to-video probing method that leverages temporal token masking and emphasizes more dynamic video regions.
Benchmarked on the public Drive\&Act dataset, \textsc{T-Mask} improves cross-view top-1 accuracy by $+1.23\%$ over strong probing baselines and $+8.0\%$ over PEFT methods, without adding any parameters.
It proves particularly effective for underrepresented secondary activities, boosting recognition by $+5.42\%$ under the trained view and $+1.36\%$ under cross-view settings.
This work provides encouraging evidence that adapting foundation models with lightweight probing methods like \textsc{T-Mask}  has strong potential in fine-grained driver observation, especially in cross-view and low-data settings. 
These results highlight the importance of temporal token selection when leveraging foundation models to build robust driver monitoring systems. 
Code and models will be made available \href{https://github.com/th-nesh/T-MASK}{here} to support ongoing research.

\end{abstract}

\begin{IEEEkeywords}
Driver Activity Recognition, Cross-view Generalization, Vision Foundation Models.
\end{IEEEkeywords}

\section{Introduction}

Changes in camera perspective present a fundamental challenge for driver monitoring systems.
Past studies have shown that the accuracy of a fully fine-tuned Convolutional Neural Network (CNN) can drop from [63\%] to near-random performance when the camera angle facing the driver shifts~\cite{drive_and_act}.
Yet most driver activity recognition frameworks are still trained and evaluated under identical viewpoint conditions~\cite{peng2022transdarc,multifuser,cm2net,roitberg2020cnn_spatialtemporal}.
Where cross-view settings are considered~\cite{drive_and_act, peng2022transdarc}, they are often treated only as minimal side experiments and consistently reveal drastic performance drops.

\begin{figure}[!t]
\vspace{-0.5em} 
    \centering
    \includegraphics[width=1\linewidth]{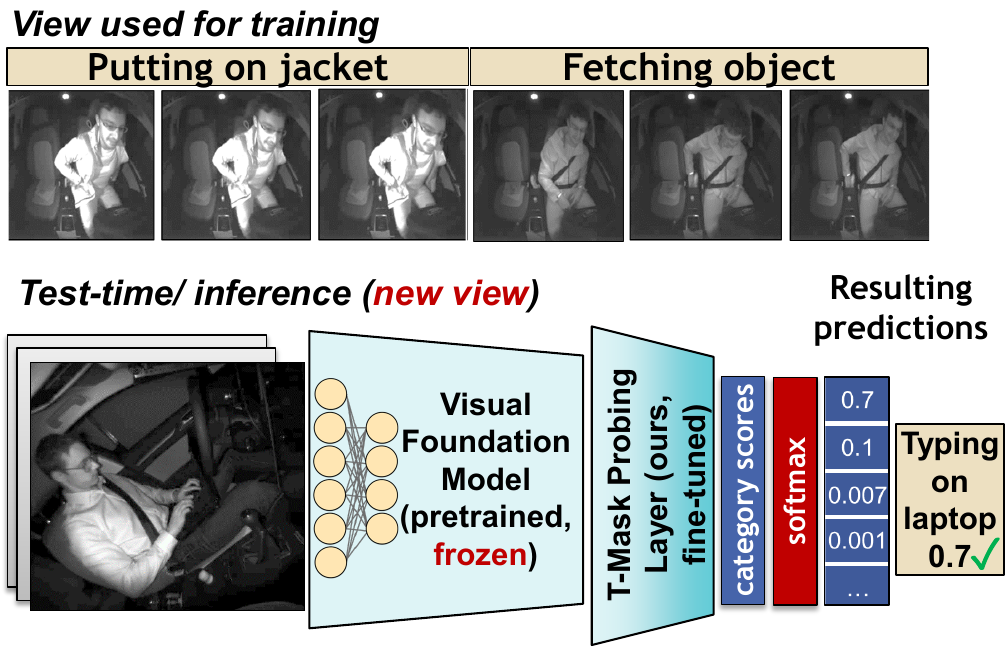}
    \caption{High-level overview of our approach for secondary driver activity recognition under unseen viewpoints.
We ask how well visual foundation models (FMs), fine-tuned with minimal parameters for in-cabin recognition, adapt to shifts in perspective. 
To address their limitations, we introduce \textsc{T-Mask}, a lightweight adaptation module that highlights dynamic, action-relevant tokens, improving robustness to viewpoint changes.}
    \label{fig:ov_teaser}
    \vspace{-1.5em} 
\end{figure}

\begin{figure*}[!t]
\vspace{-1em} 
    \centering
    \includegraphics[width=1\textwidth]{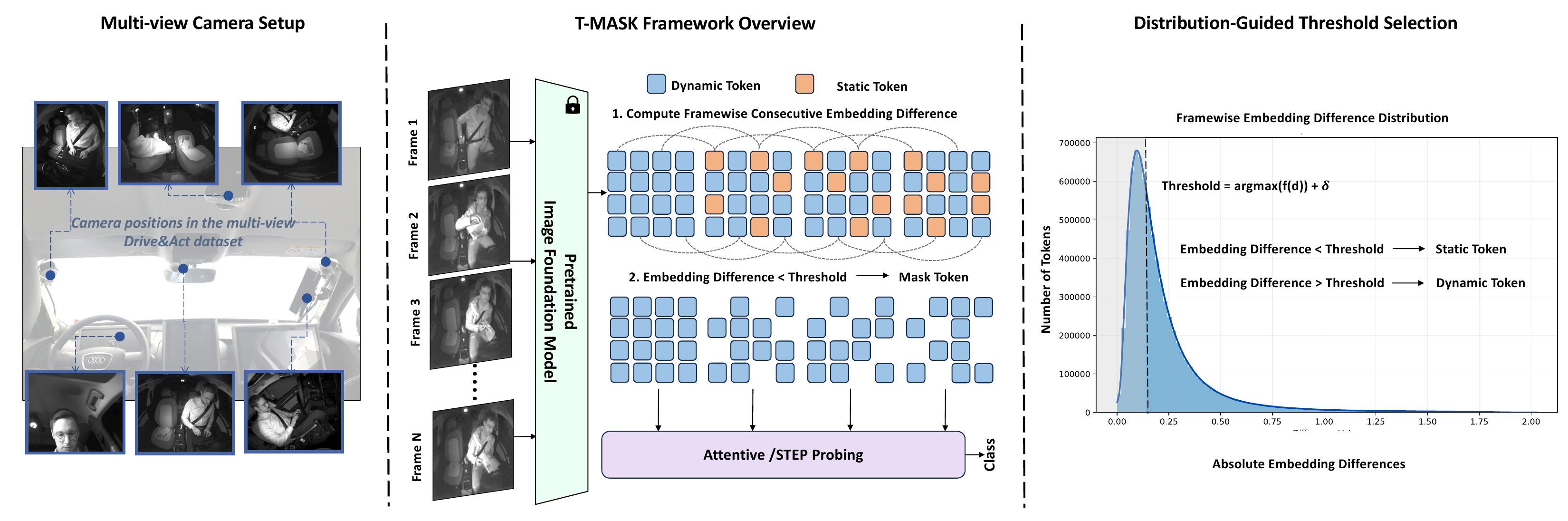}
    \caption{Overview of our proposed \textsc{T-Mask} framework for cross-view driver activity recognition. Left: Multi-view camera setup in the Drive\&Act dataset showing diverse camera viewpoints. Middle: T-MASK identifies and masks static, low-motion tokens by computing consecutive embedding differences across frames, enabling the probe to focus on dynamic, action-relevant features. Right: The masking threshold is determined via distribution-guided selection using the mode of the embedding difference distribution plus a small offset $\delta$, ensuring a principled, data-efficient approach to separating static and dynamic tokens.}
    \label{fig:teaser}
    \vspace{-1em} 
\end{figure*}


Given the limitations of general driver activity models in dealing with viewpoint variance, foundation models \cite{bommasani2021opportunities,CLIP,dinov2}, which are pre-trained on extensive and diverse datasets in a self-supervised manner, offer a promising alternative for capturing more generalized representations.
These models can be fine-tuned for various downstream tasks, setting new standards for state-of-the-art performance. 
While these models have been successful in general human activity recognition~\cite{STEP,v-jepa,aim,St_adaptor,VitaCLIP,m2_clip}, their ability to handle variations of camera perspective, especially in driving settings, remains unexplored.

Given the great success of pretrained foundation models, we ask: can they also represent fine-grained driver activities and cope with changes in camera perspective?
To investigate this, we systematically study cross-view generalization by probing two popular image foundation models (DINOv2 and CLIP) on one in-cabin camera view and evaluating them directly on unseen views.
We focus primarily on probing approaches, as they offer the most extreme form of parameter efficiency by keeping the backbone entirely frozen and only adding a minimal number of new parameters, and consider different mechanisms, ranging from simple linear heads to more advanced attention-based probing methods.
For completeness, we also benchmark against parameter-heavy PEFT~\cite{aim,St_adaptor,VitaCLIP,m2_clip} methods, which update small inserted modules or prompts inside the backbone, as well as full model fine-tuning \cite{liu2021video_swin,multifuser,tsm}.
While PEFT aims to reduce the number of trainable parameters, it still modifies a non-trivial portion of the network, making it substantially heavier than pure probing. 
We also extensively analyze cross-view performance relative to viewpoint shifts, quantifying pose differences with diverse metrics, such as geodesic \cite{geodesic} and SE(3) distances \cite{SE3}.

Building on this setup, we also introduce \textsc{T-Mask}, a new probing framework leveraging simple temporal masking aimed at suppressing static spatial features across frames, encouraging probes to focus on motion and dynamic cues – signals that are far more likely to generalize across different viewpoints.
Our experiments conducted on the public large-scale Drive\&Act dataset confirm that lightweight probing approaches, especially when augmented with \textsc{T-Mask}, are more robust to changes in camera perspective and outperform both PEFT methods and fully fine-tuned models.
\textsc{T-Mask} establishes a new state-of-the-art on Drive\&Act under cross-view and same-view regimes.

\noindent The contributions of this work are summarized as follows:
\begin{itemize}
\item 
We propose \textsc{T-Mask}, a temporal token masking strategy based on inter-frame token distances to suppress static tokens. This lightweight, plug-and-play method improves video classification performance when used with attention-based probing, without adding parameters.

    \item We analyze cross-view generalization in driver activity recognition using the Drive\&Act dataset, which features multi-view in-cabin recordings. Using the SE(3) geodesic distance between camera poses, we quantify viewpoint differences and show that recognition performance degrades with increased pose variation - highlighting the need for view-robust adaptation strategies.
    
    
    
    \item We benchmark our method against both probing and parameter-efficient fine-tuning (PEFT) baselines. Our token masking strategy improves cross-view mean accuracy by $+1 - 2$\% over probing methods, and outperforms PEFT methods by up to +8\% in mean accuracy, demonstrating superior generalization to viewpoint changes without additional parameters.
    
\end{itemize}
\section{Related Work}
\subsection{Cross-view Driver Activity Recognition}

Cross-view driver activity recognition is critical for robust in-cabin understanding in autonomous vehicles, where camera perspectives vary due to sensor placement. While general activity datasets like NTU120~\cite{liu2019ntu}, and Toyota Smart Home~\cite{dai2022toyota} support cross-view evaluation, Drive\&Act~\cite{drive_and_act} serves as a dedicated benchmark for fine-grained, multi-view driver activity recognition. 
Several methods have been proposed for general driver activity recognition using Convolutional neural networks \cite{dr_cnn1,dr_cnn3,roitberg2020cnn_spatialtemporal}, multimodal fusion \cite{mm_dr,multifuser}, and attention-based methods \cite{peng2022transdarc} in the fixed view setting. While these approaches achieve strong performance in single-view scenarios, they do not explicitly address generalization to novel viewpoints.
Both Drive\&Act\cite{drive_and_act} and TransDARC~\cite{peng2022transdarc} report significant performance drops under cross-view settings, highlighting the need for more robust approaches. Prior methods, such as Star-Transformer~\cite{ahn2023star} and PoseC3D~\cite{duan2022revisiting}, use handcrafted spatiotemporal models, while skeletal-based methods~\cite{chen2021channelwisetopologyrefinementgraph,BIAN2023103655} improve view invariance at the expense of contextual richness. Recent works like MultiFuser~\cite{multifuser}, CM2-Net~\cite{cm2net}, and MIFI~\cite{mifi2024} enhance performance through multimodal or multi-camera fusion but are computationally intensive and require multiple input streams. In contrast, our method, \textsc{T-Mask}, is the first to address cross-view generalization from a single training view. By masking static tokens based on temporal dynamics, \textsc{T-Mask} improves generalization without modifying the backbone or adding parameters, offering a lightweight and practical solution for real-world driver monitoring.

\subsection{Probing and PEFT approaches for Video Recognition}
Image foundation models like CLIP~\cite{CLIP} and DINOv2~\cite{dinov2} are increasingly adapted to video tasks through either probing or parameter-efficient fine-tuning (PEFT).
Probing methods freeze the backbone and train lightweight heads using strategies like central frame selection~\cite{CLIP}, frame averaging~\cite{dinov2}, or attention-based fusion~\cite{video_glue,v-jepa, ActionCLIP}. While efficient, most probing approaches ignore temporal order due to permutation-invariant designs~\cite{set_transformer}. Recently, Self Attention Temporal Embedding probing ~\cite{STEP} addresses this by explicitly modeling sequence dynamics, improving performance on fine-grained datasets. PEFT methods, in contrast, insert small trainable modules into frozen backbones to enable adaptation. ST-Adaptor~\cite{St_adaptor} introduces bottleneck adapters that compress and project spatiotemporal features for efficient learning. AIM~\cite{aim} adds lightweight adapters across spatial, temporal, and MLP blocks to extend temporal reasoning. VitaCLIP~\cite{VitaCLIP} uses multimodal prompts to inject temporal priors into visual features, and M2-CLIP~\cite{m2_clip} introduces a TED-Adapter to enhance temporal alignment across frames. While effective on large-scale datasets, these approaches often struggle with domain-specific, fine-grained tasks like Drive\&Act, as noted by ~\cite{STEP}. Crucially, no prior work has evaluated the cross-view generalization capabilities of probing or PEFT methods using foundation models, particularly in the context of driver activity recognition, leaving a significant gap that we address with this work.

\section{Methods}
To address the challenge of cross-view generalization in driver activity recognition, we propose \textsc{T-Mask}, a novel framework that leverages frozen vision foundation models through temporally-guided token masking. 
The key idea of  \textsc{T-Mask} is to enhance conventional  probing of foundation models  by injecting inductive bias about temporal salience into the token selection pipeline.
We detail the problem setting (Sec. \ref{sec:problem}), the architecture of our probing-based framework (Sec. \ref{sec:framework}), and introduce \textsc{T-Mask} --  a token masking strategy guided by temporal variation and statistical thresholding to enhance view-invariant representation learning (Sec. \ref{sec:tmask}). 
Finally, we provide more details about the process of distribution-based threshold selection (Sec. \ref{sec:distr}).

\subsection{Problem Statement}
\label{sec:problem}
We investigate the use of vision foundation models for driver activity recognition, with a specific emphasis on achieving robustness to viewpoint shifts -- a common challenge in real-world driver monitoring systems.
Let $\mathcal{D}_{\text{train}}$ denote the training dataset, containing videos captured from a fixed set of viewpoints $\mathcal{V}_{\text{train}}$. At test time, the model is evaluated on a dataset $\mathcal{D}_{\text{test}}$, which includes videos from both seen viewpoints $\mathcal{V}_{\text{train}}$ and novel, unseen viewpoints $\mathcal{V}_{\text{novel}}$, where $\mathcal{V}_{\text{train}} \cap \mathcal{V}_{\text{novel}} = \emptyset$. 
Given a sequence of video frames $\mathbf{x} = {x_1, \ldots, x_t}$ from a test-time viewpoint $\mathcal{V}_{\text{test}} \in \mathcal{V}_{\text{train}} \cup \mathcal{V}_{\text{novel}}$, the goal is to predict the correct action label $y$.

\subsection{Framework Overview}
\label{sec:framework}
Our framework bridges frozen image-based foundation models (e.g., DINOv2, CLIP) with the video-based task of cross-view driver activity recognition. 
 We treat the pretrained image encoder $\theta_{\text{frame}}$ as a frozen feature extractor, avoiding any backbone modification, and train a lightweight \textit{probing head} $f$ on top.
Given an input video, frame-level features are first extracted using $\theta_{\text{frame}}$ and passed into an attention-based probing module such as Attentive Probing, Self-Attention Probing, or Self-Attention Temporal Embedding Probing~\cite{STEP, set_transformer, v-jepa,ActionCLIP}. 
These probing mechanisms aggregate temporal information using lightweight attention layers, enabling temporal modeling  without modifying the backbone.

To enhance robustness under viewpoint shifts, we introduce \textsc{T-Mask}, a token masking mechanism that dynamically filters out static, view-dependent tokens with low motion across time. \textsc{T-Mask} can be used with any attention-based probing framework, encouraging  to focus on dynamic, action-relevant cues without increasing model complexity or altering the foundation model. Only the probing module $f$ is updated using labeled training data from $\mathcal{D}_{\text{train}}$, while $\theta_{\text{frame}}$ remains frozen throughout. During evaluation, we report performance separately for samples from trained-view ($\mathcal{V}_{\text{train}}$) and cross-view ($\mathcal{V}_{\text{novel}}$) test sets to assess generalization capability under unseen camera perspectives.

\subsection{Temporal Token Masking - \textsc{T-Mask}}
\label{sec:tmask}

In standard probing pipelines, all spatio-temporal patch tokens extracted by the frozen backbone are treated equally across frames. 
However, not all tokens contribute equally to action understanding — many represent static background regions that are tightly coupled to specific camera viewpoints. These static tokens introduce view-specific biases and dilute the motion dynamics essential for recognizing actions. Motivated by this, we introduce \textsc{T-Mask}, a temporal token masking mechanism designed to filter out such static, view-dependent tokens based on their motion dynamics.

Given a sequence of frame-wise patch embeddings ${\mathbf{z}_1, \mathbf{z}_2, \dots, \mathbf{z}_T}$ extracted by the frozen image encoder, we quantify the temporal variation at each token position $i$ by computing the L1 distance between consecutive frames:

\begin{equation}
d_t^{(i)} = \left\| \mathbf{z}_t^{(i)} - \mathbf{z}_{t+1}^{(i)} \right\|_1
\label{eqn:l1}
\end{equation}

The intuition of \textsc{T-Mask} is simple: tokens corresponding to dynamic, action-relevant regions (such as hands or objects being manipulated) will exhibit higher variation over time, while static tokens (e.g., car dashboard, seat) will show little to no variation.
We always retain the first frame without any masking to preserve a stable temporal reference. For subsequent frames, we apply a masking rule: if the temporal difference $d_t^{(i)}$ for a token is below a threshold $\tau$, the token is considered static and masked out using key padding masks during temporal attention. This selective masking allows the temporal probing mechanism to attend predominantly to motion-relevant features, making the learned video representation less sensitive to viewpoint-specific artifacts and more robust across views.

Importantly, \textsc{T-Mask} is \textit{architecture-agnostic}: it can be plugged into any attention-based temporal probing strategy (such as Attentive Probing, Self-Attention Probing, or Self-Attention Temporal embedding probing) without requiring any modification to the backbone model or adding significant computational overhead.

 \begin{figure}[ht!]
    \centering
\includegraphics[width=1\columnwidth]{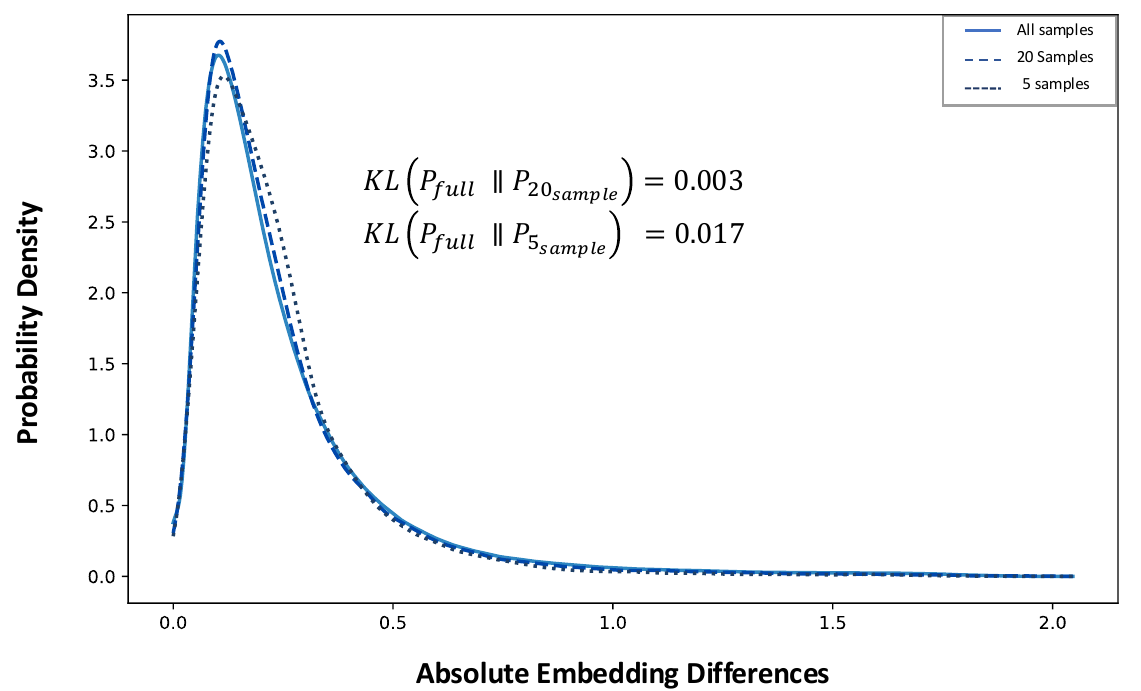}
\caption{Embedding distribution stability under subsampling. KL-divergence between the full-token embedding distribution and subsampled versions shows minimal divergence, indicating that a small number of samples is sufficient to approximate the full embedding distribution for effective token masking.}
\label{fig:prob_dist}
    \vspace{-1em} 
\end{figure}
\subsection{Distribution Guided Thresholding}
\label{sec:distr}

Setting an appropriate threshold $\tau$ for masking is crucial: a threshold that is too low would fail to suppress view-dependent noise, while one that is too high risks removing meaningful dynamic information. Rather than manually tuning $\tau$ through costly trial-and-error, we propose a distribution-guided thresholding approach that leverages the underlying statistics of the token differences.

We analyze the distribution of the L1 differences $d$ computed across all tokens and frames. As illustrated in Figure~\ref{fig:teaser}, this distribution consistently exhibits a sharp peak (mode) corresponding to static tokens, followed by a long tail representing dynamic motion. Based on this observation, we set the masking threshold as:

\begin{equation}
\tau = \arg\max f(d) + \delta
\label{eqn:tau}
\end{equation}

where $f(d)$ denotes the empirical distribution and $\delta$ is a small positive offset chosen via a lightweight hyperparameter search.
While the exact value of $\delta$ is selected experimentally, the mode-based thresholding strategy ensures that the initial guess for $\tau$ is statistically grounded, making the method robust across datasets and settings.

An important advantage of this approach is data efficiency. As shown in Figure~\ref{fig:prob_dist}, the distribution of token differences stabilizes quickly: even using only 5 or 20 sampled videos, the KL divergence from the full dataset distribution remains extremely low (0.017 and 0.003 respectively). Thus, threshold estimation does not require access to the full dataset and can be reliably performed with a small subset of videos— making the method scalable and practical for real-world applications where full datasets may not be readily available.

\section{Results and Analysis}
 \subsection{Dataset \& Implementation Details}

\noindent We conduct our experiments on the public Drive\&Act~\cite{drive_and_act} benchmark for fine-grained driver observation, featuring $34$ secondary activities from $15$ drivers. It includes NIR videos from six viewpoints, making it ideal for cross-view HAR.  We use the NIR front-top view for training and other views for evaluation, following the data splits and evaluation protocol of~\cite{drive_and_act}.
 We adopt standard evaluation protocols from video classification tasks \cite{vivit} by temporally sampling three clips from each video and averaging their logits to assess performance. Following conventions in human activity recognition \cite{aganian2023object,drive_and_act,peng2022transdarc,roitberg2020cnn_spatialtemporal,coarse_net}, we use mean class accuracy, top-1, and top-5 accuracy as our evaluation metrics. Due to the unbalanced nature of the Drive\&Act dataset, we consider balanced accuracy to be the primary evaluation metric. 
 Since the number and perspectives of views vary across datasets, we compute \textit{mean novel-view accuracy metrics} to evaluate how effectively the models generalize to different viewpoints.


\noindent\textbf{Training details:} 
We use publicly available implementations and pretrained weights for DINOv2~\cite{dinov2} and CLIP~\cite{CLIP}. The backbones are frozen, and probes are trained for 60 epochs using 4 NVIDIA A100 GPUs with a batch size of 32. For consistency, all models use 16 frames per clip.

\subsection{View Sensitivity and Cross-View Robustness}

We start by analyzing how sensitive pretrained foundation model representations are to changes in camera viewpoint in the context of driver monitoring. 
In practice, viewpoints vary not just in location but also in orientation, making cross-view generalization essential.
Standard evaluations often average results across different test cameras, but this can obscure important differences: some views are close to the training setup, others much farther.
To address this, we compute the pose difference between camera views using the \textbf{SE(3)} distance — a metric that captures both translational and rotational differences between camera poses. 
By quantifying pose dissimilarity, we can systematically study how model performance degrades with increasing viewpoint shifts and whether improvements from methods like \textsc{T-Mask} provide greater robustness as the pose gap widens.

We use a simplified approximation of the Riemannian geodesic distance on the SE(3) manifold \cite{SE3} inspired by \cite{kendall2015posenet}. This formulation jointly captures both translational and rotational discrepancies between camera poses, while remaining computationally efficient and straightforward to implement. It is defined as:
\[
\text{SE(3)} = \sqrt{\|\mathbf{t}\|^2 + ( \theta)^2}
\]
where $\|\mathbf{t}\|$ denotes the Euclidean norm of the relative translation, and $\theta$ is the \textit{geodesic distance}\cite{geodesic} between the camera orientations, capturing the shortest rotation angle on the unit sphere. In the Drive\&Act dataset, each view may correspond to multiple camera poses due to slight variations in placement. To address this, we compute the average of the extrinsic parameters across all instances for a given view, and use these mean poses to calculate the SE(3) distance. This enables a consistent and representative quantification of inter-view dissimilarity.

\begin{table}[ht!]

\centering
\caption{Per-view Top-1 accuracy and performance drop relative to the trained view (Inner Mirror), comparing the best probing mechanism (Baseline) and \textsc{T-Mask}. \textsc{T-Mask} consistently reduces the drop across views.
}
\label{tab:pose_vs_accuracy_comparison}
\resizebox{\columnwidth}{!}{
\begin{tabular}{lcccccc}
\toprule
 \textbf{View} & \textbf{SE(3)} & \multicolumn{2}{c}{\textbf{Baseline}}& \multicolumn{2}{c}{\textbf{Baseline\textsc{+ T-Mask}}}&\textbf{$\Delta$Drop $(\uparrow)$} \\

& & \textbf{Top-1}& \textbf{Drop} & \textbf{Top-1}& \textbf{Drop} & \\
\midrule
Inner Mirror (Train)     & --    & \textbf{78.55} & \textbf{0.00} & \textbf{78.96} & \textbf{0.00} & -- \\
Steering Wheel           & 1.257 & 31.02 & \textcolor{red}{-47.53} & 32.17 & \textcolor{red}{-46.79} & \textcolor{green!60!black}{\textbf{+0.74}} \\
A-Column Driver          & 1.817 & 26.97 & \textcolor{red}{-51.58} & 27.45 & \textcolor{red}{-51.51} & \textcolor{green!60!black}{\textbf{+0.07}} \\
A-Column Co-driver       & 1.864 & 24.73 & \textcolor{red}{-53.82} & 25.98 & \textcolor{red}{-52.98} & \textcolor{green!60!black}{\textbf{+0.84}} \\
Ceiling                  & 3.275 & 17.29 & \textcolor{red}{-61.26} & 19.47 & \textcolor{red}{-59.49} & \textcolor{green!60!black}{\textbf{+1.77}} \\
\bottomrule
\end{tabular}
}
\label{tab:se(3)}
\end{table}
Table \ref{tab:se(3)} presents the SE(3) distances from the trained \textit{Inner Mirror} view to all test-time views alongside their respective Top-1 accuracies and accuracy drops. A clear inverse correlation emerges: views with larger SE(3) distances (e.g., Ceiling) exhibit greater performance degradation. This validates SE(3) distance as a reliable proxy for view difficulty. More importantly, \textsc{T-Mask} consistently reduces the performance drop across all views, including the most distant one, highlighting its ability to promote geometric robustness.

\subsection{Comparison with Probing Baselines}

Next, we compare the performance of \textsc{T-Mask} against standard probing baselines such as linear probing \cite{dinov2} and attention-based probing mechanisms such as attentive probing \cite{coca,set_transformer,v-jepa}, self-attention probing \cite{ActionCLIP}, and the self-attention temporal embedding probing \cite{STEP}. 
All experiments use a frozen DinoV2 backbone.

\begin{table}[ht!]
\centering
\caption{Cross-view performance comparison of probing mechanisms with and without \textsc{T-Mask} on Drive\&Act dataset. T-Mask yields consistent gains across all baselines.}
\label{tab:Probing_baselines_comparison}
\resizebox{\columnwidth}{!}{
\begin{tabular}{lcccccc}
\toprule
\textbf{Probing} & \textbf{Pretrain} & \textbf{Method} & 
\multicolumn{3}{c}{\cellcolor{green!25}\textbf{Cross View}} \\
\cmidrule(lr){4-6}
 \textbf{Mechanism} & & & \textbf{Mean Acc.} & \textbf{Top-1 Acc.} & \textbf{Top-5 Acc.} \\
\midrule

Linear Probe \cite{dinov2} & DinoV2 & -  & 12.10 & 16.97 & 51.82 \\
\hline

& & Vanilla  & 15.35 & 21.30 & 57.52 \\
Attn Probe \cite{set_transformer,coca} & DinoV2 & \textsc{T-Mask} &  17.65 & 22.34 & 59.23 \\
& & \textbf{$\Delta$} & \textcolor{green!60!black}{\textbf{+2.30}} & \textcolor{green!60!black}{\textbf{+1.04}} & \textcolor{green!60!black}{\textbf{+1.71}} \\
\hline

& & Vanilla & 16.67 & 22.84 & 59.62 \\
Self-Attn Probe \cite{ActionCLIP} & DinoV2 & \textsc{T-Mask} & 17.85 & 22.78 & 60.58 \\
& & \textbf{$\Delta$} & \textcolor{green!60!black}{\textbf{+1.18}} & \textcolor{red}{\textbf{-0.06}} & \textcolor{green!60!black}{\textbf{+0.96}} \\
\hline

Self-Attn Temporal & & Vanilla & 19.64 & 26.28 & 57.91 \\
Embedding Probe \cite{STEP} & DinoV2 & \textsc{T-Mask} &  \textbf{20.85} & \textbf{27.51} & \textbf{61.36} \\
 && \textbf{$\Delta$}  &\textcolor{green!60!black}{\textbf{+1.21}} & \textcolor{green!60!black}{\textbf{+1.23}} & \textcolor{green!60!black}{\textbf{+3.45}} \\

\bottomrule
\end{tabular}
}
\end{table}

As shown in Table~\ref{tab:Probing_baselines_comparison}, \textsc{T-Mask} consistently improves performance across nearly all probing strategies. For instance, when applied to attention-based probing, \textsc{T-Mask} increases mean accuracy by +2.30\%, top-1 accuracy by +1.04\%, and top-5 accuracy by +1.71\%. With self-attention probing, it improves mean accuracy by +1.18\% and top-5 accuracy by +0.96\%, with a negligible drop in top-1 accuracy. The largest gains are observed when integrating \textsc{T-Mask} with Self Attention temporal embedding probing, which integrates temporal modelling into the probing mechanism, results in improvements of +1.21\% in mean accuracy, +1.23\% in top-1 accuracy, and +3.45\% in top-5 accuracy.
While the absolute improvements may appear modest, it is important to note that baseline performance in the cross-view setting is substantially lower due to viewpoint shifts. As such, even small percentage gains correspond to significant relative improvements in robustness. These results highlight that dynamically suppressing static tokens helps probes focus on informative features, improving cross-view generalization.

\subsection{View-Invariance Analysis in Embedding Space}
\begin{figure}[ht!]
    \centering
\includegraphics[width=1\columnwidth]{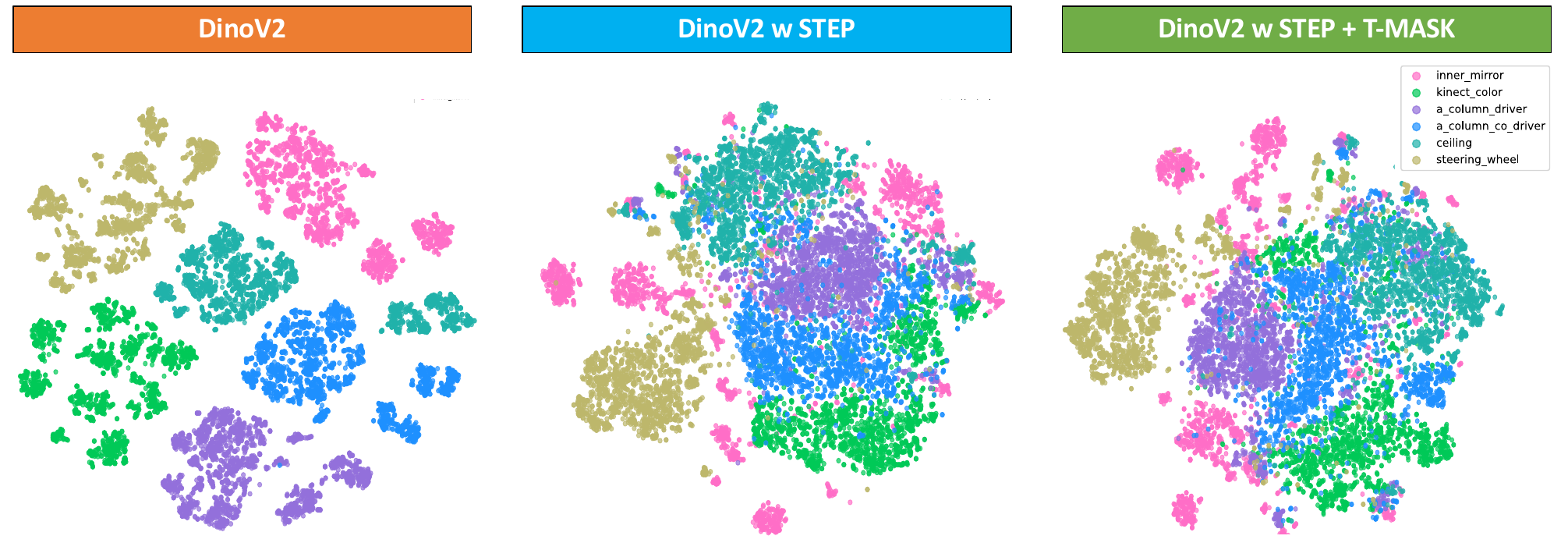}
\caption{t-SNE plots of Drive\&Act embeddings. DINOv2 shows strong view-based clustering. STEP reduces view separation. STEP + \textsc{ T-Mask} further improves overlap, enhancing view-invariance.}
\label{fig:t_sne_probe}

\end{figure}

 To better understand what changes in the representation space, we now analyze whether the learned embeddings themselves become more view-invariant by examining the structure of the learned embedding space.
 Ideally, a robust model should produce similar embeddings for the same activity, regardless of camera viewpoint. 
 To explore this, we present t-SNE visualizations of the learned feature space across six different camera views in Figure~\ref{fig:t_sne_probe}.

 \begin{figure}[ht!]
    \centering
\includegraphics[width=1\columnwidth]{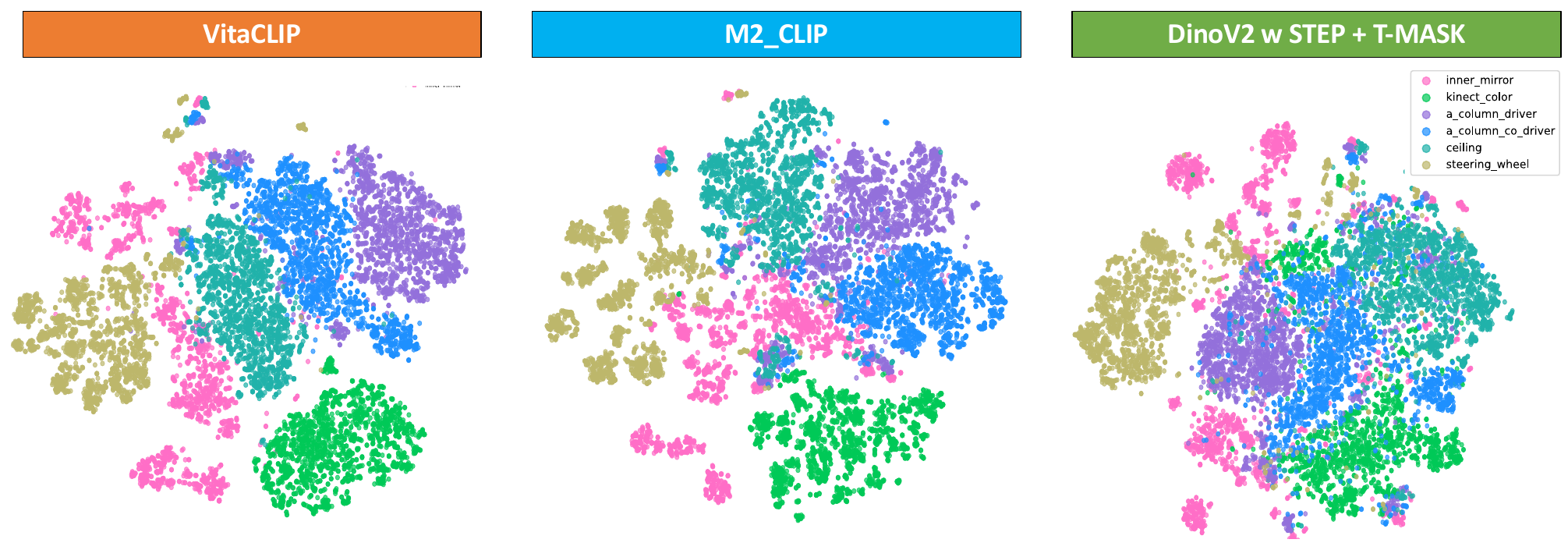}
\caption{t-SNE plots comparing PEFT methods and our probing-based approach. VitaCLIP and M2-CLIP show clear view-specific clustering, indicating poor view-invariance. In contrast, STEP + \textsc{ T-Mask}  achieves higher view overlap, demonstrating superior generalization across viewpoints.}
\label{fig:t_sne_peft}

\end{figure}

The vanilla embeddings from DINOv2 form clearly separated clusters based on viewpoint, indicating strong view-specific biases in the frozen image foundation model. Applying the STEP probing mechanism partially mitigates this separation—embeddings from different views begin to overlap more substantially. Incorporating \textsc{T-Mask} further tightens this overlap across views, suggesting that masking static tokens improves cross-view consistency. This overlap is desirable, as the goal is to cluster representations of the same action regardless of viewpoint. Hence, reduced inter-view separation signals improved view-invariance.

To broaden our comparison, we also analyze t-SNE plots from two popular PEFT methods: VitaCLIP (prompt-based) and M2-CLIP (adaptor-based). As illustrated in Figure~\ref{fig:t_sne_peft}, both methods show more prominent view separation compared to \textsc{T-Mask}, indicating that PEFT models are also susceptible to view-specific biases. This observation is consistent with our quantitative evaluations, where \textsc{T-Mask} significantly outperforms PEFT methods in cross-view recognition accuracy.

\begin{table}[ht!]
\centering
\caption{Silhouette scores (↓ lower is better) and cross-view Top-1 accuracy for various methods. \textsc{T-Mask} achieves the best view-invariance and recognition performance on Drive\&Act.}
\label{tab:silhouette_scores}
\begin{tabular}{lcc}
\toprule
\textbf{Method} & \textbf{Silhouette Score(↓)} &\textbf{Top-1 Acc.} \\
\midrule
DINOv2 \cite{dinov2}        & 0.2369 &16.97 \\
Vita-CLIP \cite{VitaCLIP}           & 0.0601 & 18.95\\
M2-CLIP \cite{m2_clip}         & 0.0580 & 24.52\\
STEP \cite{STEP}       & 0.0467 & 26.28\\
\textbf{STEP + \textsc{ T-Mask} (Ours)} & \textbf{0.0432} &\textbf{27.51} \\
\bottomrule
\end{tabular}
\end{table}

Since the degree of overlap in t-SNE plots can be difficult to quantify, we complement our qualitative findings with the Silhouette Score— a metric that evaluates how distinctly different views are clustered in the embedding space. A lower score indicates greater overlap, which in this context reflects better view-invariance. As summarized in Table~\ref{tab:silhouette_scores}, raw embeddings from DINOv2 produce a high silhouette score of 0.2369, reflecting strong view-specific separation. This score progressively drops with more effective probing strategies: VitaCLIP and M2-CLIP yield moderate reductions (0.0601 and 0.0580 respectively), while STEP (0.0467) and our proposed STEP + \textsc{T-Mask} (0.0432) achieve the lowest values, suggesting significantly reduced view bias. Notably, these silhouette scores closely correlate with cross-view Top-1 accuracy, which improves from 16.97\% for DINOv2 to 27.51\% for STEP + \textsc{ T-Mask}. This consistent trend further validates that reduced view separability translates to enhanced generalization across unseen viewpoints, confirming the effectiveness of \textsc{T-Mask} in promoting robust, view-invariant representations.

\subsection{Robustness in Recognizing Data-scarce Driver Activities}
In addition to enhancing view-invariance, we observe that \textsc{T-Mask} significantly improves recognition of underrepresented driver activities, for which only few training samples are available. 
To evaluate this, we analyze the performance of different methods across both \textit{common} and \textit{rare} driver activities, under both,  trained-view and cross-view settings.
Drive\&Act contains 34 fine-grained activity classes with a highly imbalanced distribution. Since neural networks struggle with data scarcity, we split the activities into \textit{common} (most frequent 50\%) and \textit{rare} (least frequent 50\%) behaviors based on class frequency.

\begin{table}[ht!]
\centering
\caption{Comparison of PEFT and probing methods on common vs. rare actions. \textsc{T-Mask} is highly beneficial for   rare actions.
}
\begin{minipage}{\columnwidth}
\begin{tabular}{l cccc } 
 
\toprule

\textbf{Method} & \multicolumn{2}{c}{\textbf {Trained View}}&  \multicolumn{2}{c}{\textbf{Cross View}}\\
 &  \cellcolor{green!25}\textbf{Common}&\cellcolor{blue!25}\textbf{Rare} &  \cellcolor{green!25}\textbf{Common}&\cellcolor{blue!25}\textbf{Rare} \\

\midrule
\multicolumn{5}{l}{\textbf{\textit{PEFT Frameworks}}}\\
ST-Adaptor \cite{St_adaptor}& 77.98&46.18&  26.15&8.59\\
AIM \cite{aim}& 74.31&50.00& 23.92&7.01\\
VitaCLIP \cite{VitaCLIP}& 74.07&40.45& 20.56&10.57\\
M2-CLIP \cite{m2_clip}& 81.89&50.63& 27.34&9.98\\
\midrule
\multicolumn{5}{l}{\textbf{\textit{Probing Frameworks}}}\\
STEP \cite{STEP}&\textbf{84.16} &49.36 &  28.15&14.95\\

STEP + \textsc{ T-Mask} (Ours)& 83.61&\textbf{54.78}& \textbf{29.19}&\textbf{16.31}\\
\bottomrule
\end{tabular}
\end{minipage}

\label{tab: common vs. rare}
\end{table}

Table~\ref{tab: common vs. rare} reports Top-1 accuracy for each category based on the common/rare class split defined in~\cite{roitberg2020cnn_spatialtemporal}. In the more challenging cross-view setting, STEP + \textsc{ T-Mask} surpasses STEP by +1.04\% on common actions and by +1.5\% on rare actions. Compared to PEFT methods, it shows gains of +1.85\% on common actions and between +5\% and +9\% on rare actions. 
In the trained view setting, \textsc{T-Mask} achieves performance comparable to PEFT methods on common actions—closely matching M2-CLIP (81.89\%) with a score of 83.61\%. The baseline STEP performs slightly better at 84.16\%. However, on rare actions, \textsc{T-Mask} delivers a notable boost, outperforming STEP by +5.5\% and PEFT baselines by +4\% to +14\%. These results further confirm the data-efficient nature of \textsc{T-Mask} and its strength in modeling underrepresented classes. Overall, the method demonstrates strong generalization across both rare categories and unseen viewpoints—key challenges in real-world driver activity recognition.

\subsection{Trained View and SOTA Analysis}
\begin{table*}[ht!]
\centering
\caption{Comparison of probing mechanisms, PEFT frameworks, and fully fine-tuned methods on the Drive\&Act dataset. Mean, top-1, and top-5 accuracy are reported for both trained and cross-view settings. (* indicates results reported from original papers.)}
\label{tab:summary_sota_peft_probing}
\begin{tabular}{lcccccccc}
\toprule
\textbf{Method} & \textbf{Pretrain} & \textbf{Tunable Params} & 
\multicolumn{3}{c}{\cellcolor{blue!25}\textbf{Trained View}} & 
\multicolumn{3}{c}{\cellcolor{green!25}\textbf{Cross View}} \\
\cmidrule(lr){4-6} \cmidrule(lr){7-9}
& & & \textbf{Mean Acc.} & \textbf{Top-1 Acc.}& \textbf{Top-5 Acc.} & \textbf{Mean Acc.} & \textbf{Top-1 Acc.}& \textbf{Top-5 Acc.} \\
\midrule

\multicolumn{9}{l}{\cellcolor{gray!20}\textbf{\textit{Probing Mechanisms}}} \\

STEP \cite{STEP}              & CLIP   & 2.6M & 48.19& 65.47&93.38& 16.09& 25.25& 61.35\\
STEP+ \textsc{T-Mask}  & CLIP   & 2.6M & 47.04& 65.05&92.15& 16.76& 26.09&60.89\\

STEP \cite{STEP}              & DINOv2 & 2.6M& 62.41 & 78.55 &\textbf{96.92}& 19.64 & 26.28 & 57.91\\
STEP+ \textsc{T-Mask}  & DINOv2 & 2.6M& \textbf{64.37} & \textbf{78.96} &96.61& \textbf{20.85} & \textbf{27.51}& 61.34\\

\midrule
\multicolumn{9}{l}{\cellcolor{gray!20}\textbf{\textit{PEFT Frameworks}}} \\
AIM \cite{aim}                & CLIP   & 11M      & 57.18 & 70.40& 95.74& 12.82 & 21.21& 53.73\\
ST-Adaptor \cite{St_adaptor}         & CLIP   & 7.1M    & 58.24 & 72.86 &94.76& 17.62 & 23.32& 54.36\\
M2-CLIP \cite{m2_clip}           & CLIP   & 14.8M     & 59.43 & 75.85& 96.66& 15.76 & 24.52& 55.19\\
VitaCLIP \cite{VitaCLIP}           & CLIP   & 28.6M     & 53.78 & 68.65& 94.92& 13.91 & 18.95& 49.01\\

\midrule
\multicolumn{9}{l}{\cellcolor{gray!20}\textbf{\textit{Fully Fine-Tuned Methods}}} \\
I3D \cite{carreira2017quo}*&   K400   &  12M& -& 63.64&-& -& 6.66& -\\
VideoSWIN \cite{liu2021video_swin}     &  K400& 88M& 58.98& 72.88& 93.54& 19.38& 26.44&\textbf{64.57}\\
TSM \cite{tsm}*& K400 & 24.3M& 62.72& 68.23& -& -& -&-\\
MultiFuser \cite{multifuser}*& NTU  &115M& 59.64& 72.56& -& -& -&-\\
\bottomrule
\end{tabular}

\end{table*}

While our primary focus is cross-view generalization, we also evaluate performance under the trained-view setting to ensure that improvements from \textsc{T-Mask} do not come at the cost of reduced accuracy in familiar viewpoints. This comparison allows us to assess whether T-MASK maintains or enhances performance in standard settings while improving generalization. Additionally, we benchmark \textsc{T-Mask} against state-of-the-art PEFT frameworks and fully fine-tuned methods to contextualize its performance relative to stronger but more computationally expensive baselines.

As shown in Table \ref{tab:summary_sota_peft_probing}, \textsc{T-Mask} combined with DINOv2 achieves the highest trained-view top-1 accuracy (78.96\%) across all evaluated methods, including PEFT frameworks and fully fine-tuned models, while using only 2.6M trainable parameters. This demonstrates that \textsc{T-Mask} not only supports strong cross-view generalization but also surpasses state-of-the-art methods even in the trained-view setting, where most approaches typically excel. In the cross-view evaluation, \textsc{T-Mask} delivers the best top-1 accuracy (27.51\%) and top-5 accuracy (61.34\%), outperforming probing baselines, PEFT methods such as M2-CLIP (24.52\%), and fully fine-tuned models like VideoSwin (26.44\%), despite their significantly higher computational cost. Notably, \textsc{T-Mask} maintains these gains without modifying the backbone or relying on multi-modal or multi-view inputs. These results underscore the effectiveness of \textsc{T-Mask} as a lightweight, plug-and-play strategy that enhances both in-view and cross-view performance, making it highly suitable for real-world deployment where efficiency and robustness are essential.

To further validate the generality of our approach, we apply \textsc{T-Mask} across multiple probing strategies using CLIP as the backbone, as shown in Table~\ref{tab:CLIP_probing}, and observe consistent improvements in cross-view performance. While CLIP-based models benefit from \textsc{T-Mask}, we find that DINOv2 consistently outperforms CLIP under both vanilla and \textsc{T-Mask} settings, as summarized in Table~\ref{tab:summary_sota_peft_probing}. This suggests that self-supervised features from DINOv2 transfer more effectively to fine-grained video tasks such as driver activity recognition.

\begin{table}[ht!]
\centering
\caption{Performance of probing mechanisms using CLIP as the backbone, with and without \textsc{T-Mask}.}
\label{tab:CLIP_probing}
\begin{tabular}{lccccc}
\toprule
\textbf{Probing} & \textbf{Pretrain} & \textbf{Method} & 
\multicolumn{2}{c}{\cellcolor{green!25}\textbf{Cross View}} \\
\cmidrule(lr){4-5}
 \textbf{Mechanism} & & & \textbf{Mean Acc.} & \textbf{Top-1 Acc.} \\
\midrule

Linear Probe \cite{dinov2} & CLIP & -  & 11.22 & 17.14 \\
\hline

& & Vanilla  & 16.25 & 25.27 \\
Attn Probe \cite{set_transformer,coca} & CLIP & \textsc{T-Mask} &  16.44 & 25.76  \\
\hline

& & Vanilla & 16.38 & 25.23  \\
Self-Attn Probe \cite{ActionCLIP} & CLIP & \textsc{T-Mask} & \textbf{17.41} & 25.74  \\

\hline

Self-Attn Temporal & & Vanilla & 16.09 & 25.25  \\
Embedding Probe \cite{STEP} & CLIP & \textsc{T-Mask} &  16.76 & \textbf{26.09}\\

\bottomrule
\end{tabular}
\end{table} 

We also attempted to implement PEFT methods with the DINOv2 backbone, but they did not yield stable or presentable results. Furthermore, the original M2-CLIP \cite{m2_clip}, ST-Adaptor \cite{St_adaptor}, and AIM \cite{aim} papers do not report results on DINOv2 or other self-supervised foundation models. For these reasons, we omit PEFT+DINOv2 results from our comparisons.

\begin{table}[ht!]
\centering
\caption{Effect of threshold $\tau$ on Drive\&Act performance. A value of 0.20 yields the best accuracy across views, showing \textsc{T-Mask}’s robustness to $\tau$ variations.}
\begin{minipage}{\columnwidth}
\resizebox{\columnwidth}{!}{
\begin{tabular}{l cccc } 
 
\toprule

\textbf{Threshold $\tau$}& \multicolumn{2}{c}{\cellcolor{green!25}\textbf {Trained View}}&  \multicolumn{2}{c}{\cellcolor{blue!25}\textbf{Cross View}}\\
 &  \textbf{Mean Acc.}&\textbf{Top-1 Acc.}&  \textbf{Mean Acc.}&\textbf{Top-1 Acc.}\\
 \midrule
0.10& 62.83&78.34&  19.68&25.38\\
0.15& 63.10&78.50& 18.86&25.58\\
\textbf{0.20 (Ours)}& \textbf{64.37}&\textbf{78.96}& \textbf{20.85}&\textbf{27.51}\\
0.25& 63.79&78.81& 19.65&25.96\\
 0.30& 63.03& 78.45& 19.03&26.07\\
 \toprule
\end{tabular}
}
\end{minipage}

\label{tab: threshold_ablation}
\end{table}
\begin{table}[ht!]
\centering
\caption{Ablation of masking strategies. \textsc{T-Mask} outperforms random and no masking, highlighting the value of dynamic token-informed masking.
}
\begin{minipage}{\columnwidth}
\resizebox{\columnwidth}{!}{
\begin{tabular}{l ccccc } 
 
\toprule

\textbf{Masking}&\textbf{Mask} &\multicolumn{2}{c}{\cellcolor{green!25}\textbf {Trained View}}&  \multicolumn{2}{c}{\cellcolor{blue!25}\textbf{Cross View}}\\
 & \textbf{Percentage}& \textbf{Mean Acc.}&\textbf{Top-1 Acc.}&  \textbf{Mean Acc.}&\textbf{Top-1 Acc.}\\
 \midrule
None& 0 \%&62.41 &78.55 &  19.64&26.28\\
Random &20\%& 61.71&78.14& 19.30&25.80\\

\textbf{\textsc{T-Mask} (Ours)}&18.96\%& \textbf{64.37}&\textbf{78.96}& \textbf{20.85}&\textbf{27.51}\\

\bottomrule
\end{tabular}
}
\end{minipage}

\label{tab: masking_ablation}
    \vspace{-1em} 
\end{table}
\subsection{Ablation Studies}

\noindent\textbf{Impact of Threshold $\tau$.}
We evaluate the sensitivity of \textsc{T-Mask} to the threshold \(\tau\), which identifies static tokens across frames. 
As defined in Equation~\ref{eqn:tau}, \(\tau\) is computed as the sum of the mode of the token difference distribution f(D) and an offset \(\delta\) to account for temporal variation noise.
 From the observed distribution, \(\arg\max(f(D))\) occurs at approximately 0.10; sweeping \(\delta\) from 0 to 0.2 reveals that \(\delta\)=0.10 (i.e., \(\tau=0.20\) yields the best performance.
As shown in Table~\ref{tab: threshold_ablation}, \(\tau=0.20\) achieves the highest mean and top-1 accuracy across trained and cross-view settings. Nearby values also perform well, indicating that \textsc{T-Mask} is robust to variations in \(\tau\).


\noindent\textbf{Random vs Distance-Based Masking.} 
To further validate the importance of dynamic, token-informed masking, we compare \textsc{T-Mask} with random and no-masking baselines. Random masking matches \textsc{T-Mask}’s 20\% mask ratio for fairness in comparison.
As shown in Table~\ref{tab: masking_ablation}, \textsc{T-Mask} outperforms both baselines,while random masking even reduces accuracy compared to no masking. This highlights the importance of selectively removing static tokens to enhance cross-view robustness and fine-grained recognition.


\vspace{-1em}
\section{Conclusion}
We studied the challenge of cross-view generalization in driver activity recognition -- how well models trained from one camera view perform when tested on others. 
To understand this, we quantified pose differences between camera views and analyzed view sensitivity both through accuracy trends and embedding space structure.
We also proposed \textsc{T-Mask}, a lightweight, plug-and-play probing strategy that enhances cross-view driver activity recognition by masking static tokens based on temporal dynamics. Without modifying the backbone or adding parameters, \textsc{T-Mask}  improves robustness to unseen viewpoints and underrepresented actions. Experiments on Drive\&Act show that smart temporal token selection can outperform both PEFT and fully fine-tuned methods, setting new state-of-art results in same-view and cross-view settings.


\vspace{-0.5em}

\section*{Acknowledgment}
The research published in this article is supported by the Deutsche Forschungsgemeinschaft
(DFG) under Germany’s Excellence Strategy – EXC 2120/1 –390831618. 
The authors also gratefully acknowledge the computing time provided on the high-performance computer HoreKa by the National High-Performance Computing Center at KIT. 
HoreKa is partly funded by the German Research Foundation (DFG).

{\small
\bibliographystyle{IEEEtran}
\bibliography{IEEEabrv,IEEEexample}

\end{document}